\documentclass[sigplan]{acmart} 
\renewcommand\footnotetextcopyrightpermission[1]{} 
\usepackage{booktabs}   
\usepackage{subcaption} 

\usepackage{enumitem}

\usepackage[linesnumbered,ruled,vlined]{algorithm2e}
\usepackage{xcolor}

\begin{document}

\title[GraphGen+]{GraphGen+: Advancing Distributed Subgraph Generation and Graph Learning On Industrial Graphs}

\author{Yue Jin, Yongchao Liu, Chuntao Hong}
\affiliation{%
  \institution{Ant Group, China}
  \city{}
  \country{}}
\email{{jinyue.jy, yongchao.ly, chuntao.hct}@antgroup.com}

\maketitle
\pagestyle{plain}
Graph-based computations are crucial in a wide range of applications, where graphs can scale to trillions of edges. To enable efficient training on such large graphs, mini-batch subgraph sampling is commonly used, which allows training without loading the entire graph into memory. However, existing solutions face significant trade-offs: online subgraph generation, as seen in frameworks like DGL and PyG, is limited to a single machine, resulting in severe performance bottlenecks, while offline precomputed subgraphs, as in GraphGen, improve sampling efficiency but introduce large storage overhead and high I/O costs during training. To address these challenges, we propose \textbf{GraphGen+}, an integrated framework that synchronizes distributed subgraph generation with in-memory graph learning, eliminating the need for external storage while significantly improving efficiency. GraphGen+ achieves a \textbf{27$\times$} speedup in subgraph generation compared to conventional SQL-like methods and a \textbf{1.3$\times$} speedup over GraphGen, supporting training on 1 million nodes per iteration and removing the overhead associated with precomputed subgraphs, making it a scalable and practical solution for industry-scale graph learning.

\section{Introduction}
Graph-based computations are essential for many applications.
As real-world graphs expand in size, encompassing billions or even trillions of edges~\cite{ching2015one}, efficiently processing and learning from these massive graphs has become increasingly important.
To enable efficient training on such large graphs, mini-batch subgraph sampling is commonly used, which allows training without loading the entire graph into memory. However, existing solutions face significant trade-offs.
Frameworks like DGL~\cite{soleymani2021deep} and PyG~\cite{fey2019fast} are designed for online subgraph generation, where subgraphs are sampled dynamically during training, but limited to single-machine environments, making them inefficient for large-scale graphs. The sampling and training processes are constrained by the memory and computational capabilities of a single machine, leading to high overhead and poor scalability.
Therefore, they can only handle graphs with up to a billion edges.

In contrast, frameworks such as AGL~\cite{zhang2020agl} and GraphGen~\cite{GraphGen} utilize MapReduce~\cite{dean2008mapreduce} for offline subgraph generation, significantly improving sampling efficiency. By distributing the subgraph extraction process across multiple machines, GraphGen addresses the computational bottleneck present in single-machine frameworks. However, this approach still faces a major limitation: it requires substantial storage to precompute and store the subgraphs. This is especially problematic for large graphs, as storing these precomputed subgraphs demands significant disk space and incurs high I/O costs. Furthermore, the need to read and write subgraphs from local or network disk storage introduces significant delays, slowing down the training process.
On the other hand, AGL utilizes a node-centric MapReduce paradigm, which serially processes neighbor collection when high-degree nodes occur, creating performance bottlenecks.

To overcome these limitations, we propose \textbf{GraphGen+}, a novel high-performance framework that integrates distributed subgraph generation with in-memory graph learning.
By synchronizing subgraph generation and in-memory training, GraphGen+ eliminates the need for pre-computed subgraphs and external storage, improving both storage efficiency and system performance.
Moreover, GraphGen+ employs an edge-centric MapReduce approach, enabling parallel neighbor collection even in the presence of high-degree nodes.
The experimental results show that GraphGen+ achieves a 27$\times$ speed increase in subgraph generation compared to traditional SQL-like methods and a 1.3$\times$ speedup over GraphGen. In addition, it supports training at 1 million nodes per iteration, making it a scalable and practical solution for large-scale graph learning tasks.

In summary, we present GraphGen+, a high-performance framework that is oriented to industry-scale graph learning: \textcircled{1} Integrates distributed subgraph generation with in-memory graph learning, enabling high-performance graph learning on large-scale graphs. \textcircled{2} Optimizes subgraph distribution among workers using a load balancing strategy, ensuring efficient parallel execution and minimizing resource under utilization. \textcircled{3} Reduces storage overhead by eliminating the need for external storage, improving both speed and scalability. \textcircled{4} Achieves a 27× speedup in subgraph generation compared to traditional SQL-like methods and a 1.3× speedup over GraphGen, making it a scalable and practical solution for large-scale graph learning tasks.

\section{Methods}

GraphGen+ integrates distributed subgraph generation with in-memory graph learning to optimize large-scale graph processing. The method consists of four main steps:

\textbf{(1) Graph Partitioning:}  
The input graph $\mathcal{G} = (\mathcal{V}, \mathcal{E})$ is partitioned between multiple worker nodes to allow parallel computation and reduce memory overhead on individual workers. This step is performed by a \textbf{coordinator node}, which is responsible for managing the data distribution. The partitioning strategy aims to minimize cross-worker communication while ensuring that each worker receives a balanced portion of the graph.  

\textbf{(2) Load-Balanced Subgraph Mapping:}  
To evenly distribute the computational workload between workers, the \textbf{coordinator node} constructs a \textit{balance table} that maps seed nodes to worker memory. The seed nodes are assigned sequentially to the workers in a round-robin fashion. If the total number of seed nodes is not a multiple of the number of workers, any remaining seed nodes are \textbf{discarded} to maintain the balance of the workload. This design ensures that each worker processes an approximately equal number of subgraphs, preventing any single worker from becoming a bottleneck.  

\textbf{(3) Distributed Subgraph Generation:}  
Each worker extracts subgraphs by collecting the edges associated with its assigned seed nodes. If an edge $E = (v_1, v_2) \in \mathcal{E}$ belongs to multiple seed nodes, it is \textbf{replicated} across multiple workers to ensure the completeness of all subgraphs.  

To mitigate the computational overhead caused by high-degree (hot) nodes, we employ a \textbf{tree reduction strategy}. Instead of having all workers communicate directly with a central aggregator, we organized them into a hierarchical tree structure. Each non-leaf worker \textbf{partially processes} and aggregates its assigned subgraphs before passing the results to its parent node in the tree. This step continues recursively until the final aggregation is performed on the root worker. By distributing computation across multiple levels, this method reduces communication overhead, balances workload more effectively, and minimizes bottlenecks caused by highly connected nodes. However, the effectiveness of this approach depends on the bandwidth of the network and the location of the worker, which we plan to further optimize in future work.  

\textbf{(4) In-Memory Graph Learning:}  
Once subgraphs are generated, they are immediately loaded into memory for training, eliminating storage and I/O overhead. The training process is conducted over multiple \textbf{epochs}, where each worker:  
(1) iterates over its assigned subgraphs,  
(2) performs local training updates, and  
(3) synchronizes gradients across all workers using an \textit{AllReduce} operation to ensure model consistency.  

Furthermore, \textbf{subgraph generation and training are executed concurrently}: as new subgraphs are generated, they are directly loaded into memory and used for training, enabling an efficient data pipeline. This strategy ensures that the model can learn continuously from dynamically generated subgraphs without delays.  

The entire workflow is summarized in Algorithm~\ref{algo_1}, which details the steps of partitioning, mapping, extraction, and distributed training.  

\begin{algorithm}[h]
\small
\SetAlgoLined
\KwIn{Graph $\mathcal{G} = (\mathcal{V}, \mathcal{E})$ with vertices $\mathcal{V}$ and edges $\mathcal{E}$, \\
      Seed nodes $\mathcal{S} = \{s_1, s_2, \dots, s_n\}$, \\
      Worker nodes $\mathcal{W} = \{w_1, w_2, \dots, w_m\}$, \\
      Maximum training epochs $maxEpochs$}
\KwOut{Trained graph model with optimized subgraph generation and workload distribution}

\textbf{Step 1: Graph Partitioning} \\
Distribute $\mathcal{G}$ across $\mathcal{W}$ using a distributed partitioning strategy\;

\textbf{Step 2: Load-Balanced Subgraph Mapping} \\
Shuffle $\mathcal{S}$ randomly to avoid sequential bias\; 
Initialize iterator $\textit{it} \gets$ begin of $\mathcal{S}$\;
Compute $\textit{max\_i} \gets \lfloor |\mathcal{S}| / |\mathcal{W}| \rfloor \times |\mathcal{W}|$\;
\For{$i \gets 0$ \KwTo $\textit{max\_i} - 1$}{
    \If{$\textit{it} =$ end of $\mathcal{S}$}{
        \textbf{break}\;
    }
    Assign $\mathcal{M}[\textit{it}] \gets \mathcal{W}[i \bmod |\mathcal{W}|]$\;
    Increment $\textit{it}$\;
}

\textbf{Step 3: Distributed Subgraph Generation} \\
\ForEach{edge $E = (v_1, v_2) \in \mathcal{E}$}{
    Identify all seed nodes $S \in \mathcal{S}$ where $v_1 \in S$ or $v_2 \in S$\;
    \ForEach{seed node $S$ that contains $E$}{
        Assign worker $W = \mathcal{M}[S]$\;
        Append $E$ to subgraph $Graph(S)$ on worker $W$\;
    }
}

\textbf{Step 4: In-Memory Graph Learning} \\
\For{$epoch \gets 1$ \KwTo $maxEpochs$}{  
    \While{training is not complete (e.g., loss < threshold or maxEpochs reached)}{ 
        \ForEach{worker $W \in \mathcal{W}$ in parallel}{
            Fetch next available subgraph $Graph(S)$ from memory\;
            Train model on $Graph(S)$ using mini-batch gradient descent\;
            Synchronize gradients across workers using AllReduce\;
        }
    }
}

\caption{GraphGen+ Workflow}
\label{algo_1}
\end{algorithm}

\section{Preliminary Results}
We evaluate GraphGen+ on a graph with 530 million nodes and 5 billion edges, using a GCN~\cite{kipf2016semi} model for mini-batch training.
Our experiments are conducted on a 256-node Docker cluster, where each container is allocated 8 CPU cores and 16GB of memory. For subgraph sampling, we use a 2-hop neighborhood expansion strategy, selecting 40 neighbors in the first hop and 20 neighbors in the second hop for each seed node.
Subgraph generation is completed in 3 minutes, processing 5.9 million nodes per second, which represents a 27$\times$ speedup over traditional SQL-like methods and 1.3$\times$ speedup over GraphGen.

The 1.3$\times$ speedup is primarily attributed to the Load-Balanced Subgraph Mapping, which ensures balanced workload among workers, and the Tree-Reduction technique that addresses hot node load issues.
Our system is capable of training on up to 1 million nodes per iteration, a limit derived from experimental settings that showcase efficient concurrent subgraph generation and training at this scale.

\section{Conclusion}
GraphGen+ integrates distributed subgraph generation with in-memory graph learning, achieving a \textbf{27$\times$} speedup in subgraph generation and eliminating the need for external storage.
Our approach is implemented in our graph intelligent computing system~\cite{liu2023graphtheta} for in production environment and offers a scalable, efficient solution for industrial-scale graph learning, making it well-suited for real-world applications.

\bibliographystyle{ACM-Reference-Format}
\bibliography{bibfile}

\end{document}